\documentclass[conference]{IEEEtran}
\IEEEoverridecommandlockouts
\usepackage{cite}
\usepackage{amsmath,amssymb,amsfonts}
\usepackage{algorithmic}
\usepackage{graphicx}
\usepackage{textcomp}
\usepackage{xcolor}
\usepackage[acronym]{glossaries}
\usepackage{subcaption}
\usepackage{listings}
\usepackage{adjustbox}
\usepackage{array}
\usepackage{hyperref}
\usepackage{multirow}

\def\BibTeX{{\rm B\kern-.05em{\sc i\kern-.025em b}\kern-.08em
    T\kern-.1667em\lower.7ex\hbox{E}\kern-.125emX}}

\glsdisablehyper
\newacronym{ADSE}{ADSE}{Automatic Design Space Exploration}
\newacronym{AER}{AER}{Address Event Representation}
\newacronym{AI}{AI}{Artificial Intelligence}
\newacronym{ANN}{ANN}{Artificial Neural Network}
\newacronym{API}{API}{Application Programming Interface}
\newacronym{ASIC}{ASIC}{Application-Specific Integrated Circuit}
\newacronym{AX}{AX}{Adaptive eXperimentation}
\newacronym{BPTT}{BPTT}{Back-Propagation Through Time}
\newacronym{BRAM}{BRAM}{Block RAM}
\newacronym{CNN}{CNN}{Convolutional Neural Network}
\newacronym{CPU}{CPU}{Central Processing Unit}
\newacronym{CSNN}{CSNN}{Convolutional Spiking Neural Network}
\newacronym{CU}{CU}{Control Unit}
\newacronym{DNN}{DNN}{Deep Neural Network}
\newacronym{DP}{DP}{Data Path}
\newacronym{DRAM}{DRAM}{Dynamic Random Access Memory}
\newacronym{DSE}{DSE}{Design Space Exploration}
\newacronym{DVS}{DVS}{Dynamic Vision Sensor}
\newacronym{ELUT}{ELUT}{Equivalent Look Up Table}
\newacronym{ETH}{ETH}{Eidgenössische Technische Hochschule}
\newacronym{FC}{FC}{Fully-Connected}
\newacronym{FCR}{FC-R}{Fully-Connected Recurrent}
\newacronym{FF}{FF}{Feed Forward}
\newacronym{FFFC}{FF-FC}{Feed-Forward Fully-Connected}
\newacronym{FPGA}{FPGA}{Field Programmable Gate Array}
\newacronym{FSM}{FSM}{Finite State Machine}
\newacronym{GPGPU}{GPGPU}{General Purpose Graphic Processing Unit}
\newacronym{GPU}{GPU}{Graphic Processing Unit}
\newacronym{HDL}{HDL}{Hardware Description Language}
\newacronym{IF}{IF}{Integrate and Fire}
\newacronym{INI}{INI}{Institute of Neuro-Informatics}
\newacronym{IOT}{IoT}{Internet of Things}
\newacronym{LIF}{LIF}{Leaky Integrate and Fire}
\newacronym{LUT}{LUT}{Look Up Table}
\newacronym{MAC}{MAC}{Multiply and Accumulate}
\newacronym{ML}{ML}{Machine Learning}
\newacronym{NAS}{NAS}{Network Architecture Search}
\newacronym{NE}{NE}{Network Evaluator}
\newacronym{NG}{NG}{Network Generator}
\newacronym{NLP}{NLP}{Natural Language Processing}
\newacronym{NN}{NN}{Neural Network}
\newacronym{PU}{PU}{Processing Unit}
\newacronym{RAM}{RAM}{Random Access Memory}
\newacronym{RCR}{RC-R}{Randomly-Connected Recurrent}
\newacronym{RL}{RL}{Reinforcement Learning}
\newacronym{RNN}{RNN}{Recurrent Neural Network}
\newacronym{ROM}{ROM}{Read Only Memory}
\newacronym{RSNN}{RSNN}{Recurrent Spiking Neural Network}
\newacronym{RTL}{RTL}{Register Transfer Level}
\newacronym{SBS}{SbS}{Spike-by-Spike}
\newacronym{SCNN}{SCNN}{Spiking Convolutional Neural Networks}
\newacronym{SHD}{SHD}{Spiking Heidelberg Digits}
\newacronym{SIMD}{SIMD}{Single Instruction Multiple Data}
\newacronym{SNN}{SNN}{Spiking Neural Network}
\newacronym{SOC}{SoC}{System on Chip}
\newacronym{SOA}{SoA}{State of Art}
\newacronym{SRAM}{SRAM}{Static Random Access Memory}
\newacronym{SRM}{SRM}{Spike Response Model}
\newacronym{STDP}{STDP}{Spike-Timing-Dependent Plasticity}
\newacronym{TPU}{TPU}{Tensor Processing Unit}
\newacronym{UCB}{UCB}{Upper Confidence Bound}
\newacronym{VHDL}{VHDL}{VHSIC Hardware Description Language}
\newacronym{WTA}{WTA}{Winner Takes All}

\newcolumntype{?}{!{\vrule width 1pt}}

\definecolor{myCyan}{rgb}{0.18, 0.46, 0.72}

\newcommand{\spkexp}{SpikeExplorer}

\begin{document}

\title{\spkexp{}: hardware-oriented Design Space Exploration for Spiking Neural Networks on FPGA
}

\author{
    Dario Padovano$^1$,
    Alessio Carpegna$^1$,
    Alessandro Savino$^1$,
    Stefano Di Carlo$^1$ \\
    
    $^1$Politecnico di Torino
}

\maketitle

\begin{abstract}
One of today's main concerns is to bring Artificial Intelligence power to embedded systems for edge applications. The hardware resources and power consumption required by state-of-the-art models are incompatible with the constrained environments observed in edge systems, such as IoT nodes and wearable devices. Spiking Neural Networks (SNNs) can represent a solution in this sense: inspired by neuroscience, they reach unparalleled power and resource efficiency when run on dedicated hardware accelerators. However, when designing such accelerators, the amount of choices that can be taken is huge. This paper presents SpikExplorer, a modular and flexible Python tool for hardware-oriented Automatic Design Space Exploration to automate the configuration of FPGA accelerators for SNNs. Using Bayesian optimizations, SpikerExplorer enables hardware-centric multi-objective optimization, supporting factors such as accuracy, area, latency, power, and various combinations during the exploration process. The tool searches the optimal network architecture, neuron model, and internal and training parameters, trying to reach the desired constraints imposed by the user. It allows for a straightforward network configuration, providing the full set of explored points for the user to pick the trade-off that best fits the needs. The potential of SpikExplorer is showcased using three benchmark datasets. It reaches 95.8\% accuracy on the MNIST dataset, with a power consumption of 180mW/image and a latency of 0.12 ms/image, making it a powerful tool for automatically optimizing SNNs.
\end{abstract}

\begin{IEEEkeywords}
Neuromorphic; Spiking Neural Networks; Hardware Accelerators; FPGA; Design Space Exploration; Network Architecture Search; Hyperparameter Optimization
\end{IEEEkeywords}

\section{Introduction}
\label{sec:introduction}

\glspl{SNN} represent a new computing paradigm, shaped by neuroscience models exploring networks of biological neurons \cite{maass_networks_1997}. Differently from other types of \glspl{ANN}, \glspl{SNN} mimic the behavior of biological neurons more faithfully, trying to reach the extreme energy efficiency observed in our brain. Although this goal is still far, \glspl{SNN} are already able to outperform \gls{SOA} \gls{ANN} models in many different applications, and in particular in those for which the energy consumption is somewhat constrained \cite{narayanan_spinalflow_2020}. 
\glspl{SNN} become particularly interesting when implemented through dedicated hardware co-processors. Indeed, the intrinsic efficiency of these models makes them especially suitable to be implemented on digital \glspl{ASIC} \cite{basu_spiking_2022}, \glspl{FPGA} \cite{isik_survey_2023}, and analog dedicated circuits \cite{musisi-nkambwe_viability_2021}.

In this context, one of the main challenges is how to configure the \gls{SNN} to fit the target application best: there are many different neuron models with varying degrees of biological plausibility and computing efficiency; a single model has a lot of internal parameters to tune; the network architecture itself can be modified depending on the task to perform. A manual selection of all these hyper-parameters can be very complex and could bring a non-optimal solution. At the same time, an exhaustive search for the best configuration would require too much time, given the search space size. \gls{ADSE} can represent a solution. However, while the literature is rich in works about \gls{ADSE} in the field of \glspl{CNN} \cite{wang_efficient_2022-1, ghaffari_cnn2gate_2020}  and other \gls{ANN} models \cite{czako_automaticai_2021}, this is not true for \glspl{SNN}. The few existing works on the topic focus on a single-objective optimization directed towards the improvement of the accuracy \cite{balaji_neuroxplorer_2021} or concentrate the search on a particular aspect of the network, like the input data encoding \cite{abderrahmane_design_2020}, keeping the neuron model and the network architecture fixed.

This paper presents \spkexp{}, a flexible Hardware-Oriented \gls{ADSE} framework to automatically optimize digital hardware accelerators for \glspl{SNN}, targeting \gls{FPGA} implementations. The tool supports multi-objective \gls{ADSE} driven by power consumption, latency, area, and accuracy. \spkexp{} can be specialized for whatever neuron model and hardware implementation, allowing to easily customize \gls{SNN} co-processors depending on the user requirements. This can help leverage the benefits of \glspl{SNN} in power and resource-constrained edge applications, strongly simplifying the configuration and tuning of this new network in various everyday problems. 

\section{Background}
\label{sec:background}

\gls{AI} is reaching unparalleled performance, matching human capabilities in complex tasks like pattern recognition, \gls{NLP}, and object detection. However, it still stands orders of magnitude behind human intelligence regarding energy efficiency \cite{samsi_words_2023}. When it comes to optimization,  nature excels, and its solution to minimize brain power consumption is to make neurons communicate through asynchronous sequences of spikes. \glspl{SNN} are based on the same communication approach, drawing inspiration from biology to model how neurons react to these spikes. In a \gls{SNN}, the information is encoded in the timing of the spikes, regarding them as binary events. Therefore, from a computational perspective, neurons in an \gls{SNN} handle streams of single-bit data, strongly reducing the overall required complexity. Different neuron models can react to spikes in different ways. Neurons can be interconnected differently, and training algorithms can tune the resulting network on a specific problem. This leads to a large design space that requires proper techniques to be analyzed and reduced. The following sections show a subset of all the possible design choices that can be considered in the search.

    \begin{figure}[t]
         \centering
         \includegraphics[width=\columnwidth]{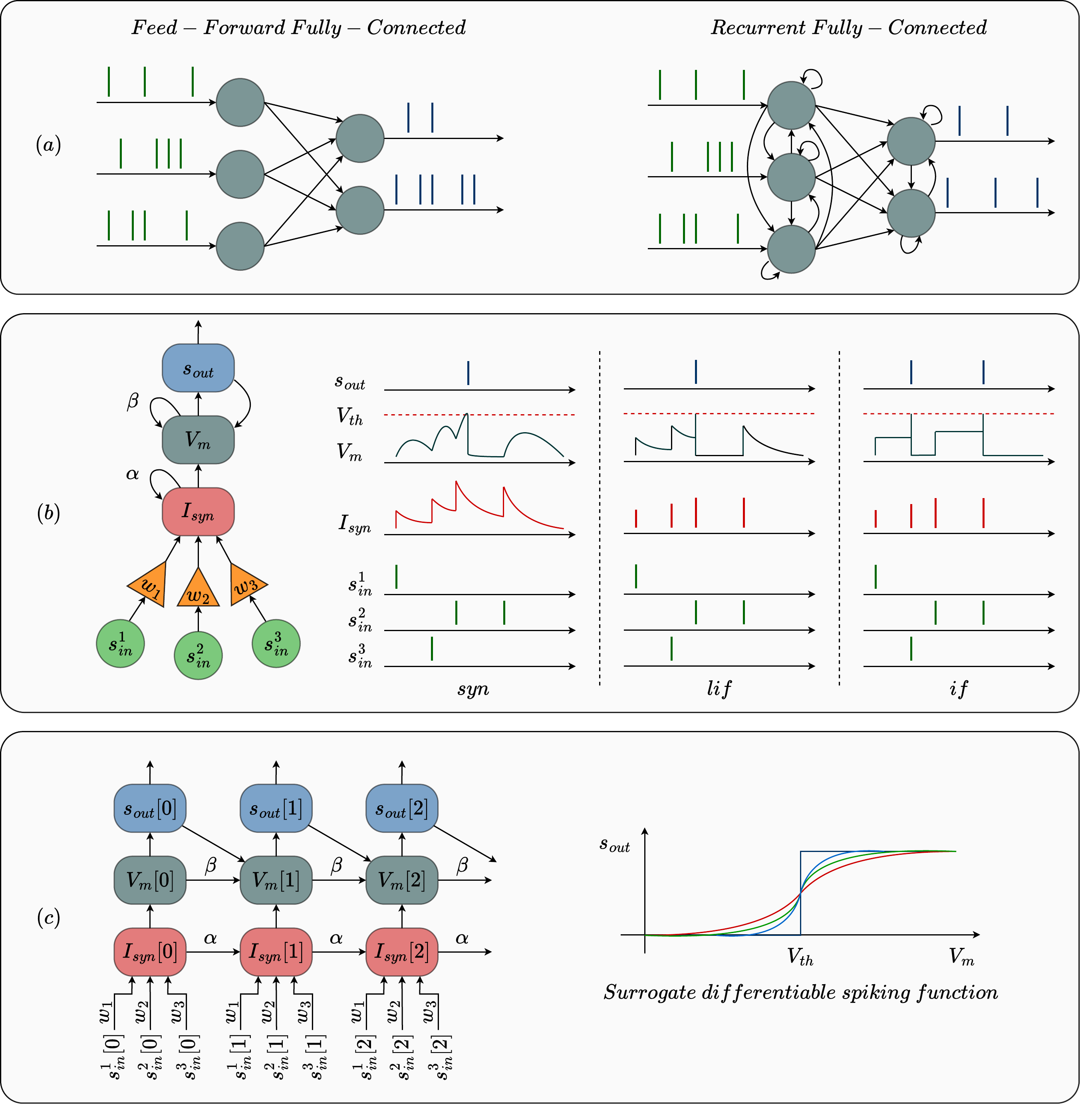}
         \caption{\gls{SNN} Design Space: (a) different network architectures; (b) different neuron models; (c) different surrogate functions and learning parameters}
         \label{fig:design_space}
    \end{figure}

\subsection{Network architecture}
\label{subsec:snn_architecture}

Neurons can be interconnected in various patterns to construct complex networks. One widely used connection scheme is the \gls{FC} architecture, which can extract complex features from input data. Neural network connections typically adhere to either a \gls{FF} architecture that facilitates a linear information flow from inputs to outputs or adopt recurrent structures with feedback connections, allowing information to loop back. \autoref{fig:design_space}\textcolor{myCyan}{a} shows the two alternative architectures. Spiking neurons inherently exhibit recurrence since their state is computed starting from the previous one. Hence, the architecture retains information from previous states even in the context of \gls{FF} \glspl{SNN}. However, explicit feedback connections might be necessary to capture longer dependencies or complex dynamics. \spkexp{} aims to optimize  \gls{FC} architectures organized in layers, interconnected both in a \gls{FF} and recurrent manner. These architectures are general enough to address most \gls{ML} problems.

\subsection{Neuron models}
\label{subsec:neuron_models}

The first computational models of biological neurons were developed starting from the accurate observation of electrical propagation inside neural cells \cite{hodgkin_quantitative_1952}. Nonetheless, for practical computational tasks, such a high level of biological fidelity is unnecessary and overly complex. Different simplified alternatives have been developed in the last decades \cite{izhikevich_simple_2003}. The most used one is the family of \gls{IF} models \cite{brunel_quantitative_2007}, able to describe neuron dynamics with limited computational complexity. Essentially, an \gls{IF} neuron functions as an integrator, accumulating spikes over time and subsequently fires itself a spike if the cumulative value surpasses a predefined threshold. Inputs are transmitted to the neuron via synapses, where they undergo pre-processing before reaching their destination. The most complex \gls{IF} neuron model is the conductance-based \gls{LIF} model, described, in discrete time, by Equations \ref{eq:synapse}, \ref{eq:neuron} and \ref{eq:fire}. The synaptic current generated by the synapse has a dynamic response to the input spikes. The $i-th$ synapse weights the input spikes through its synaptic weight $w_i$. Without stimuli, the current decays exponentially toward a rest value. \autoref{eq:synapse} shows a compact description of the total net synaptic current $I_{syn}$ received by the neuron: each synapse has its weight and the product $W \cdot s_{in}[n]$ represents the weighting operation performed by N synapses on as many inputs. All the synapses share the exponential decay rate $\alpha$ (with $\alpha$ < 1).

    \begin{equation}
        I_{syn}[n] = \alpha \cdot I_{syn}[n-1] + W \cdot s_{in}[n] 
        \label{eq:synapse}
    \end{equation}

The input current is then integrated by the neuron into its membrane potential $V_m$. If the result stays below a threshold value $V_{th}$, $V_m$ follows a temporal dynamic similar to the synaptic one, so it decays exponentially with a decay rate $\beta$ (with $\beta < 1$). If instead $V_m$ exceeds $V_{th}$, it is reset by the function $R$, and an output spike $s_{out}$ is generated. 

    \begin{equation}
        V_m[n] = \beta \cdot (V_m[n-1] - s_{out}[n-1] \cdot R[n]) + I_{syn}[n]
        \label{eq:neuron}
    \end{equation}

\autoref{eq:fire} shows two possible mechanisms for the reset operation. In the first case, called hard reset, $V_m$ is always reset to zero when the threshold is exceeded, i.e., when a spike is generated. In the subtractive reset alternative, the threshold is subtracted by $V_m$.

    \begin{align}
		\begin{aligned}
    		R_{hard}[n] = V_m[n-1] \\
    		R_{sub}[n] = V_{th}
	    \end{aligned}
     \quad
		&
		s_{out}[n] = \begin{cases}
			1, & \mbox{if } V_m > V_{th} \\
			0, & \mbox{if } V_m \leq V_{th}
		\end{cases}
		\label{eq:fire}
	\end{align}

In the rest of the paper, what was just described will be called the synaptic model (abbreviated as \textit{syn}), following the terminology used in \cite{eshraghian_training_2021}.
The conductance-based \gls{LIF} model can be simplified by removing the dynamic response of the synapse, considering only the synaptic weight, as it is generally done in \glspl{ANN}. This is equivalent to set $\alpha=0$ in \autoref{eq:synapse}. The result is a simple \gls{LIF} model, referenced in the rest of the paper as \textit{lif}. Finally, the neuron's dynamic response could also be neglected, transforming the neuron into a simple integrator. This can be obtained by setting $\beta=1$. The result is a basic \gls{IF} model (\textit{if)}. 
\autoref{fig:design_space}\textcolor{myCyan}{b} summarizes these three behaviors, showing an example of their temporal response to spikes.
Therefore, even considering only the \gls{IF} family of models, it is clear that a lot of knobs can be tuned during design. For example, $\alpha$ and $\beta$ determine how fast the exponential decay is, influencing the capability of the neuron to keep the memory of past information, while $V_{th}$ and $R$ affect the firing rate of the neuron and the timing of the output spikes. 

\subsection{Training}
\label{subsec:training}

Training \glspl{SNN} remains an active area of research, drawing inspiration from both biological observations \cite{markram_spike-timing-dependent_2012} and classical supervised approaches in \gls{ML}. However, a significant challenge arises when using supervised approaches with \glspl{SNN} due to the non-differentiable nature of the output spike function $s_{out}$ concerning the neuron's state $V_m$, as illustrated in \autoref{eq:neuron}. Consequently, the traditional backpropagation training algorithm \cite{rumelhart_learning_1986} and its derivatives are impractical. To address this issue, a commonly adopted approach involves substituting the spike function with a differentiable surrogate during the backward pass \cite{neftci_surrogate_2019}. Various options exist for this surrogate, typically encompassing smoothed versions of the step function, such as the sigmoid and its derivatives, arc or hyperbolic tangents, among others. Once the non-differentiability is mitigated, training proceeds akin to that of \glspl{RNN}: the network can be unrolled and trained using the \gls{BPTT} algorithm, propagating the output error across both space (layers of the network) and time (unrolled states of the network). \autoref{fig:design_space}\textcolor{myCyan}{c} shows the unrolling process together with a graphical example of surrogate spike functions used during the backward pass. Subsequently, selecting and fine-tuning an appropriate surrogate function and traditional backpropagation parameters, including learning rate, regularization parameters, optimizer settings, etc., are essential steps in the training process.

\subsection{Automatic Design Space Exploration}
\label{subsec:adse}

When working with complex systems such as \glspl{SNN}, the numerous degrees of freedom make a comprehensive exploration of the design space impractical. This challenge is compounded when crafting specific hardware implementations, where synthesizing and simulating architectures can consume significant time. Over the past few decades, researchers have sought ways to optimize and speed up the search for optimal architectures in electronic systems, a pursuit intensified by the proliferation of \gls{AI} and \glspl{ANN}. One approach to reducing the search space involves randomly selecting a subset of points and focusing exploration solely on them. Despite its simplicity, this can prove effective in many cases \cite{marti_optimization_2020}. Nevertheless, structured and systematic alternatives abound in the literature, many drawing inspiration from biological evolution, like evolutionary and genetic algorithms \cite{ferrandi_multi-objective_2008}, Simulated Annealing and Extremal Optimization \cite{savino_redo_2018}, and \gls{RL} \cite{saeedi_design_2023}.
Among optimization techniques, Bayesian optimization \cite{reagen_case_2017} stands out as a robust solution, particularly for its ability to converge rapidly even with complex models. It effectively addresses the exploration-exploitation dilemma \cite{march_exploration_1991}, balancing exploring new solutions and exploiting known ones. Instead of directly interacting with the real objective function (e.g., accuracy of the network), which might be computationally expensive to evaluate, Bayesian optimization builds a simpler, approximate model. It is typically based on Gaussian processes or other probabilistic models. Initially, this surrogate model makes some assumptions about the objective function based on the limited information available. As more data points are collected through evaluations of the actual objective function, the surrogate model becomes refined and better approximates the true function. A Bayesian optimizer relies on an acquisition function, considering both exploration and exploitation aspects to decide which point in the search space to evaluate next. Exploration involves trying out points in the search space that are uncertain or have not been explored much to gain more information about the objective function and potentially discover better solutions. Moreover, exploitation involves focusing on areas of the search space likely to yield good results based on the current knowledge provided by the surrogate model. The acquisition function balances these two aspects to guide the search effectively. Thanks to this methodical approach, Bayesian optimization demonstrates efficacy in converging to solutions, even in scenarios involving numerous parameters in the search, rendering it a valuable tool for optimizing \glspl{SNN}.

\section{Related works}
\label{sec:related}

Among the large plethora of \glspl{ANN} models, \glspl{SNN} are the ones that mostly require dedicated hardware co-processors. Indeed \glspl{SNN} are characterized by high computational parallelism, lightweight communication channels exchanging asynchronous spikes, and co-location of memory and computing. This does not fit well with the Von-Neumann computing paradigm adopted in general-purpose computers, which relies on a limited number of computational units exchanging data and instructions with a centralized memory. Even specialized architectures, such as \glspl{GPU} and \glspl{TPU}, optimized for standard \gls{ANN} workloads, struggle to process \glspl{SNN} efficiently. Consequently, employing dedicated neuromorphic hardware emerges as the most efficient solution, especially in contexts where efficiency is the primary concern.

In this landscape, one of the solutions that is gaining attention is to exploit the reconfigurability of \glspl{FPGA} to design application-specific \gls{FPGA}-based \glspl{SNN} co-processors \cite {han_hardware_2020, gupta_fpga_2020, li_fast_2021, carpegna_spiker_2022, carpegna_spiker_2024}. The advantage of using \glspl{FPGA} is their intrinsic reconfigurability, which reduces design time and makes network customization easier. This enables the fine-tuning of hardware implementations for \glspl{SNN} according to specific problem requirements, configuring the hardware to deploy the most optimized solution.
An automatic optimization for \gls{SNN} architectures is critical when considering these dedicated hardware implementations, particularly for resource-constrained edge applications. In such scenarios, the optimization of the network targets multiple objectives: together with the fine-tuning of the model on a specific problem, minimizing power consumption, area occupancy, and latency become integral parts of the optimization goals.

 Within this framework, tools are available to support \gls{FPGA} hardware designs. For instance, $E^3NE$ \cite{gerlinghoff_e3ne_2022} provides a library of elementary blocks to build \gls{RTL} descriptions of \gls{SNN} architectures. On the other hand, Spiker+ \cite{carpegna_spiker_2024} provides a framework to automatize the generation of the \gls{SNN} \gls{RTL} models starting from a high-level network description, providing a library of possible models and network architectures.
However, a crucial gap remains: given the availability of various neuron blocks and architectures, how can the network be optimized to achieve the highest possible accuracy while constraining other metrics such as latency, power consumption, or area? Some works on \gls{NAS} for \glspl{SNN} exist. For example, authors in \cite{kim_neural_2022} perform a single-objective search to optimize \gls{SNN} accuracy across different benchmark applications. A hardware-oriented exploration is conducted in \cite{balaji_neuroxplorer_2021} to optimize networks for deployment on available neuromorphic platforms. Finally, the first example of \gls{FPGA}-oriented \gls{DSE} is found in \cite{abderrahmane_design_2020}. The work focuses on the best encoding technique for input data translation into spike sequences. What is still missing is a comprehensive tool that, leveraging power, latency, and area estimations of specific hardware blocks, conducts a multi-objective optimization and provides an architectural description ready for integration onto an \gls{FPGA}.

\section{Materials and Methods}
\label{sec:methods}

    \begin{figure}[t]
		\centering
		\includegraphics[width=\columnwidth]{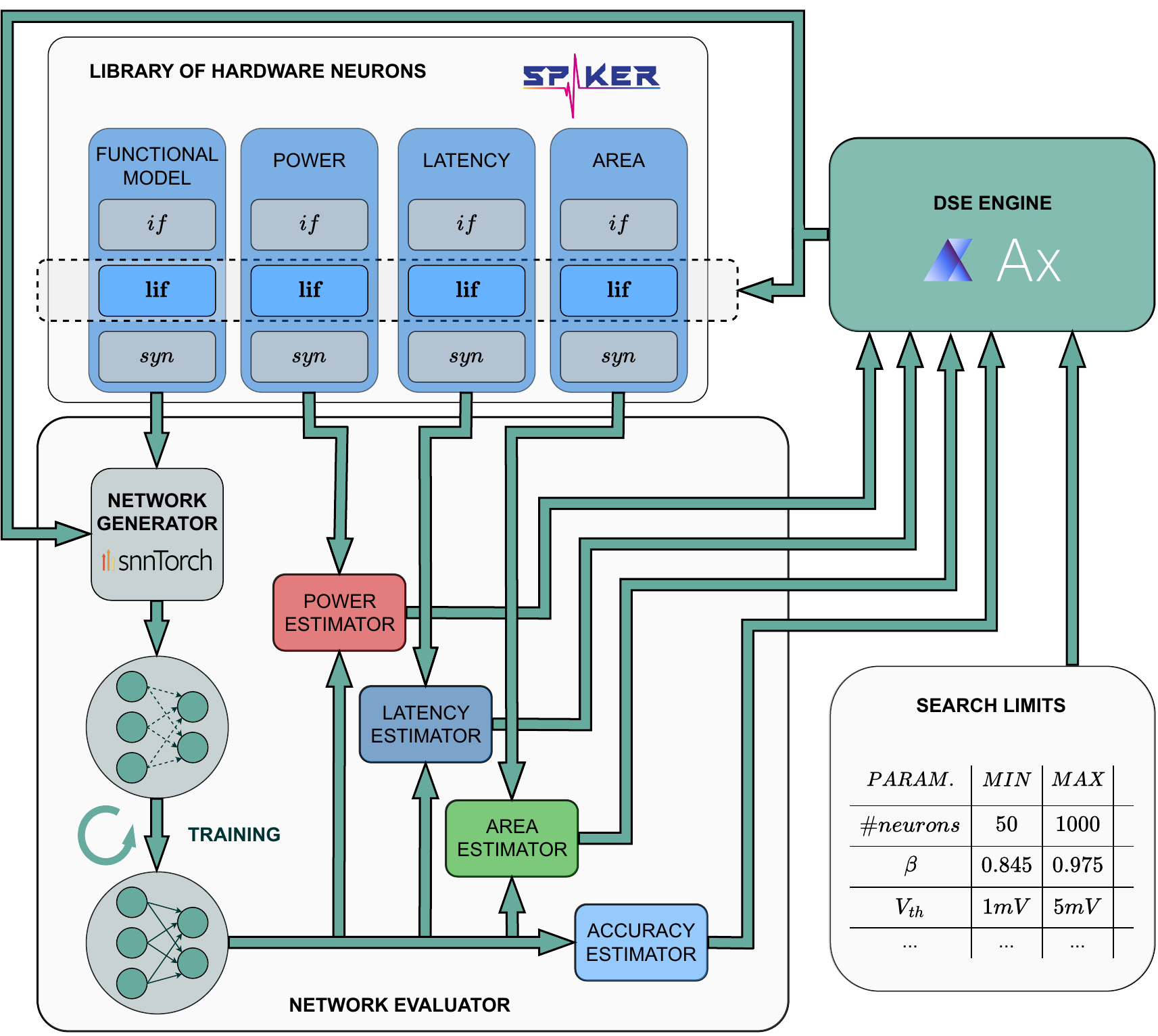}
		\caption{\spkexp{} general architectures, including (i) a library of hardware neurons, (ii) a network evaluator estimating the performance of selected implementations, and (iii) a Bayesian \gls{DSE} engine.}
		\label{fig:spkexp}
	\end{figure}

\spkexp{} has been designed as a modular Python tool with different components connected in a closed loop. \autoref{fig:spkexp} shows a high-level view of the complete framework.

The \gls{DSE} engine is the core of the optimization framework. It aims at finding the optimal \gls{SNN} architecture and its related parameters for a given problem within a user-defined design space. The user imposes constraints by specifying which parameters must remain fixed and which require optimization. An infinite search space risks prolonged search duration and potential converging failure. Hence, users are prompted to define search limits for each optimized parameter. This ensures that the search remains bounded and manageable. For instance, limits can be set on the maximum network size, considering the available hardware resources on the target platform.

The optimization process follows a multi-objective Bayesian approach. The user can select a set of optimization targets: accuracy, area, latency, and power. While \autoref{fig:spkexp} illustrates an exploration encompassing all four potential metrics, the optimization can focus solely on a subset or even just one in the extreme case.
Once the optimization objectives are defined, the \gls{DSE} engine constructs a surrogate model for each of them and starts an iterative optimization process. The surrogate models determine the next point to explore at each iteration, aiming to optimize all required metrics. A point within the search space is defined by a set of values associated with the parameters used for the optimization.

For each explored design option, the specific \gls{SNN} architecture and configuration is forwarded to the \gls{NE}, responsible for the network construction, training, and performance evaluation. This, in turn, requires providing a training dataset. This block closes the loop by providing the \gls{DSE} engine with the characterization of the selected observation points in terms of accuracy, area, latency, and power required to update the internal surrogate models. This task requires comprehensively characterizing the various neuron models computing their individual area occupancy, power consumption, and latency. 
\spkexp{} has been intentionally designed to be versatile and compatible with any user-defined neuron characterization model. However, this paper utilizes a comprehensive characterization library derived from open-source experiments, leveraging the SPIKER+ framework \cite{carpegna_spiker_2024}.
Given the framework's complexity, the next sections overview each component separately.

\subsection{Network Generator and hardware neurons}
\label{subsec:net_eval}

The \gls{NG} is the submodule of the \gls{NE} block in charge of building the \gls{SNN} network models required for performance evaluations. It requires a set of functional models of the available neurons that can be incorporated into the network architecture. Each functional model must be characterizable for the considered optimization targets. For instance, if the optimization target is area minimization, the user must provide a characterization detailing the area occupation of each considered neuron model. This facilitates fine-tuning the search process with specific neuron implementations, which will be integrated into the customized \gls{SNN} co-processor on \gls{FPGA}.

In its current implementation, \spkexp{} supports a set of default neuron functional models based on the \gls{IF} variants described in \autoref{subsec:neuron_models}. From a functional point of view, the models are defined using the snnTorch framework \cite{eshraghian_training_2021}. This facilitates the creation of a range of networks suitable for various problems where \gls{IF} models are applicable. snnTorch enables to model and approximate the hardware neuron behavior without a precise knowledge or description of all internal details. 

Each available neuron model is associated with hardware-related information obtained using the open-source hardware models provided by the Spiker+ framework \cite{carpegna_spiker_2024}. These models are synthesized on a Xilinx XC7Z020 reference \gls{FPGA} board, and the corresponding performance metrics are extracted, such as area, power, and latency.
The following sections outline the techniques applied to characterize the default \spkexp{} neuron models. This presentation aims to provide insight into the available estimates and to explain how neurons can be characterized to fine-tune the search on a specific implementation.

\subsection{Area}
\label{subsec:area}

    \begin{figure*}[t]
		\centering
        \includegraphics[width=0.9\textwidth]{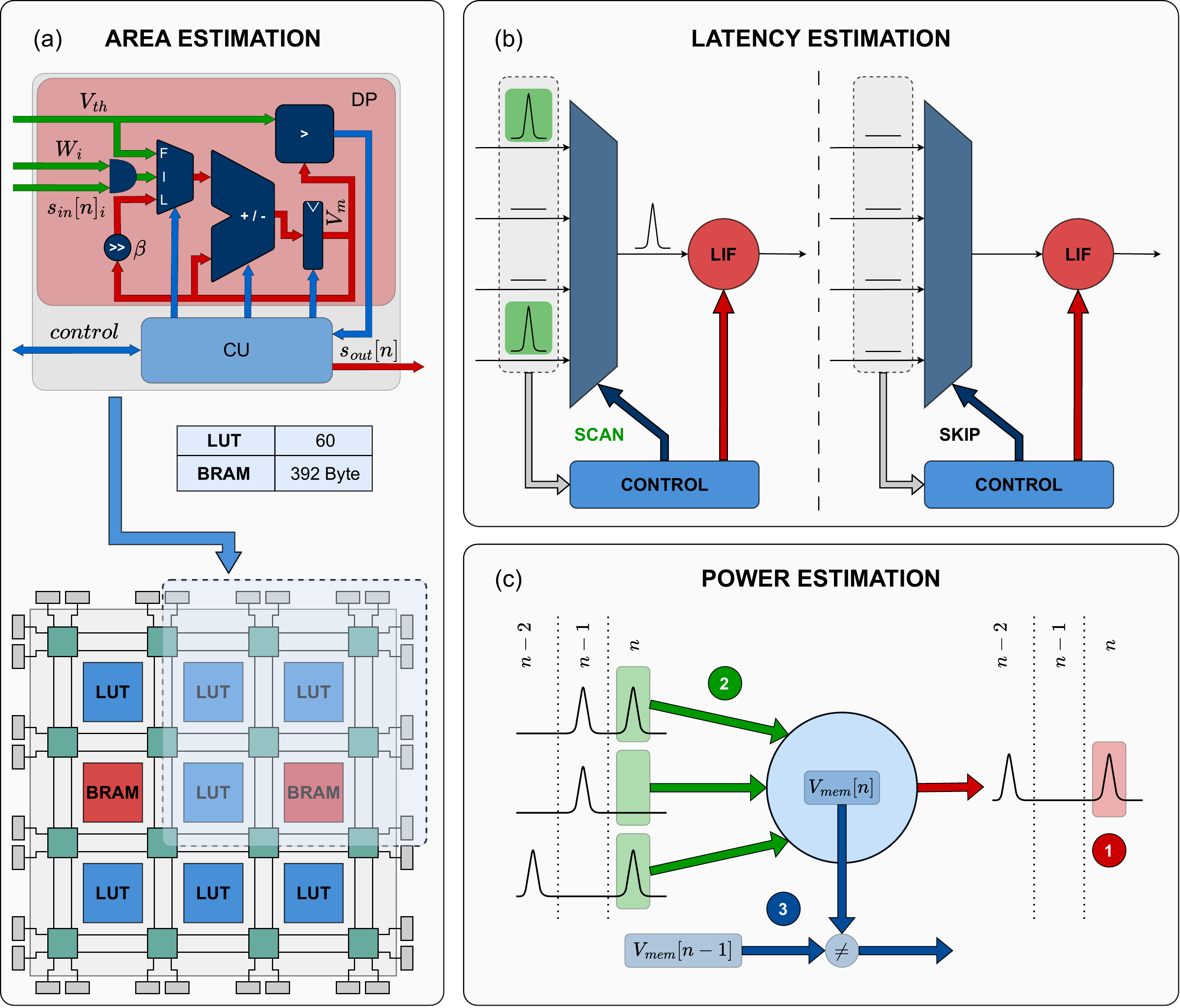}
		\caption{Metrics estimation}
		\label{fig:metrics}
	\end{figure*}

The neuron area estimation consists of two primary components: (i) the area occupied by the computational elements and (ii) the memory utilized by the synaptic connections. Both are estimated through hardware synthesis of available implementations.
\autoref{fig:metrics}\textcolor{myCyan}{a} shows an example of the synthesis of a simple \gls{LIF} model, reporting the corresponding \gls{LUT} count and the amount of memory required by synaptic weights, in this case, stored in \gls{FPGA} \gls{BRAM}. 

Quantization is a well-known technique exploited to reduce the memory footprint of \glspl{SNN} models, and several quantization frameworks exist \cite{putra_q-spinn_2021, li_quantization_2022, castagnetti_trainable_2023}. To avoid overlap with existing solutions, the default neuron library provided by \spkexp{} does not aim at optimizing quantization, focusing on higher-level architectural optimizations. Therefore, the default neuron characterization uses 32-bit data representations. Experimental results later show that this data representation preserves full-precision accuracy without introducing bias from precision reductions across different models. Although the resulting architecture may appear oversized, what truly influences the optimization process is the relative dimensions of the neurons. 
Nevertheless, the user can enlarge the library of available neurons, including architectures with different quantization levels, to drive the search toward smaller neurons with more aggressive quantization.

The total number of weights is computed at run-time after defining the \gls{SNN} architecture and integrated into the area estimation to account for different architectural structures and their impact on the area. This allows us to consider the diverse memory footprints of different architectural choices. For instance, in fully connected architectures with identical neuron count on each layer, a deeper network featuring smaller layers will incorporate fewer synaptic weights, thus necessitating less memory.
Moreover, recurrent architectures introduce an area overhead due to \gls{FC} recurrent connections that can be computed according to  \autoref{fig:memory_overhead}. 

\begin{align}
    R = \frac{Recurrent\ layer\ area}{\gls{FF}\ layer\ area} = \\
    \frac{N_{in} \cdot N_{neurons} + N_{neurons} \cdot N_{neurons}}{N_{in} \cdot N_{neurons}} = \\
    \frac{N_{in}  + N_{neurons}}{N_{in}}
    \label{fig:memory_overhead}
\end{align}

Here, $N_{in}$ denotes the number of inputs, and $N_{neurons}$ signifies the number of neurons within a specific layer.

\spkexp{} measures the overall area occupancy in terms of \glspl{ELUT} count:

    \begin{equation}
        N_{ELUT} = \sum_{l=0}^{N_{layers}} N_{neurons}^l \cdot \left( N_{LUT32} + N_{in}^l \cdot r \right)
    \end{equation}

Where  $l$ denotes the layer index, $N_{layers}$ the total number of layers, $N_{ELUT}$ represents the total number of \glspl{ELUT} occupied by the network, $N_{LUT32}$ is the number of \glspl{LUT} required by a single neuron with a 32-bit synthesis, and 

    \begin{equation}
        r =  \begin{cases}
                R, & \mbox{if l is recurrent}      \\
                1, & \mbox{if l is \gls{FF}}
        \end{cases}
    \end{equation}

For clarity in visualization, the distinction between \gls{FF} and recurrent architectures is expressed using the neuron model nomenclature. The default supported models encompass \textit{if}, \textit{rif}, \textit{lif}, \textit{rlif}, \textit{syn}, and \textit{rsyn}, where the prefix \textit{r} signifies a recurrent architecture.

\subsection{Accuracy and latency}
\label{subsec:acc_latency}

Since quantization is not the primary focus of \spkexp{}, the accuracy estimation of various network configurations used to drive the \gls{DSE} process is based on full-precision 64-bit floating-point software models constructed by the \gls{NG} using the snnTorch framework. These estimations are crucial for guiding the optimization process but should not be regarded as precise accuracy measurements for the target hardware co-processor. They represent an upper bound on the final accuracy that depends on the quantization applied when deploying the model on a real \gls{FPGA}.

In terms of latency, a clock-driven reference model is considered. In particular, \spkexp{} implements two different latency estimation models: a fixed latency model, in which each neuron is characterized by a single latency value, independent of the spiking activity, which accounts for the time required to integrate spikes and to decay or reset the neuron; and an optimized latency model, following a computational methodology like that described in \cite{carpegna_spiker_2024}. In this case, two latency values are considered: a high latency occurs when at least one input spike is present, prompting neurons to scan all inputs to identify the active ones, and a low value occurs without spikes, where the scanning process is omitted. \autoref{fig:metrics}\textcolor{myCyan}{b} shows the two considered cases. Since all the inputs are processed sequentially, the largest the number of inputs to a neuron, the highest will be the latency of that neuron in case it receives input spikes. The computational process is considered entirely parallel, making the overall latency independent of the overall number of neurons.
The approach is highly tailored to fully parallel clock-driven implementations. Alternatively, an activity-based methodology resembling the one utilized for power consumption (refer to \autoref{subsec:power}) could be adopted to accommodate event-driven approaches.

\subsection{Power}
\label{subsec:power}

The neurons' power consumption generally depends on their activity levels. This holds for clock-driven architectures, as evidenced in \cite{carpegna_spiker_2024}, and is even more pronounced in event-driven alternatives. To understand how \spkexp{} estimates the overall power consumption, it is convenient to analyze the operations involved in updating a \gls{LIF} neuron. \autoref{eq:power} shows the mathematical operations involved, obtained by merging \autoref{eq:synapse} and \autoref{eq:neuron}, setting $\alpha = 0$ and re-ordering the terms. 

    \begin{equation}
        V_m[n] = \underbrace{\beta \cdot V_m[n-1]}_\text{(3) Leak} + \underbrace{W \cdot s_{in}[n]}_\text{(2) Integrate} - \underbrace{\beta \cdot s_{out}[n-1] \cdot R[n]}_\text{(1) Fire} 
        \label{eq:power}
    \end{equation}

As the name of the model suggests, the neuron executes three main operations: \textit{Leakage} (3), \textit{Integration} (2), and \textit{Firing} (1). The equation defines the evolution of the membrane potential in its discrete-time form. \spkexp{} examines the state of each neuron at every time step $n$ to evaluate the instantaneous power consumption. It expects a characterization of the power consumed by the neuron when executing each of the reported operations. The overall power consumption is then computed by averaging the instantaneous values over the entire sequence of time steps. To accomplish this task, \spkexp{} must monitor (i) the presence of an output spike, (ii) the presence of input spikes, and (iii) the value of the state variables that change dynamically during the network operations. With \gls{LIF} and \gls{IF} models, the only state variable involved is $V_m$, while with a synaptic model, $I_{syn}$ is monitored as well. Using these, \spkexp{} understands the current state of the neuron and infers the relative consumed power, as illustrated in \autoref{fig:metrics}\textcolor{myCyan}{c}.

Observing the neuron's output reveals whether the neuron has "fired" a spike. If a spike is generated due to the threshold potential being exceeded, the membrane is reset; this is associated with a first power contribution. Inspecting the inputs, if spikes are present, they are weighted and integrated into the membrane potential, implying an additional power contribution. Eventually, without spikes, the membrane decays toward its rest value, consuming extra power. This condition happens when the membrane potential at time step $n$ differs from that at time step $n-1$. This approach facilitates a highly adaptable evaluation. For instance, in clock-driven update policies, decay consumes power at every time step, which can be factored into the Leak contribution.
Conversely, in an event-driven approach, computations occur solely in the presence of input spikes, potentially resulting in zero power consumption for the leak term. In this case, the decay power can be merged into the "integrate" term. Alternatively, a custom functional model can be used for the neuron, in which the membrane is updated only when input spikes are received. Here, by checking whether the membrane has changed value, the decay contribution can be considered only in the presence of input stimuli. Lastly, depending on factors like recent resetting or reaching asymptotic decay values due to finite precision platforms, the neuron may remain in a constant state without necessitating significant power-consuming updates.

\subsection{DSE engine}
\label{subsec:dse_engine}

    \begin{table}[h]
        \caption{Set of specifications that the user can provide}
        \label{tab:dse_params}
        \centering
        \begin{tabular}{?c?c?c|c?}

            \cline{2-4}
            
            \cline{2-4}
            
            \multicolumn{1}{c?}{}                               &
            \textbf{Parameter}                                  &
            \multicolumn{2}{c?}{\textbf{Values}}                \\

            \hline

            \hline

            \multirow{4}{*}{\rotatebox[origin=c]{90}{Net}}      &

            \# layers                                           &
            \multicolumn{2}{c?}{Discrete}                       \\

            \cline{2-4}

            \cline{2-4}

            {}                                                  &
            \# neurons/layer                                    &
            \multicolumn{2}{c?}{Set}                            \\
            
            \cline{2-4}

            \cline{2-4}

            {}                                                  &
            Architecture                                        &
            \begin{tabular}{c} 
                Feed        \\
                Forward
            \end{tabular}                                       &
            Recurrent                                           \\

            \cline{0-1}

            \cline{0-1}

            \cline{1-4}

            \multirow{7}{*}{\rotatebox[origin=c]{90}{Neuron}}   &

            Model                                               &
            \begin{tabular}{c} 
                if  \\
                lif \\
                syn \\
            \end{tabular}                                       &
            \begin{tabular}{c} 
                rif  \\
                rlif \\
                rsyn \\
            \end{tabular}                                       \\

            \cline{2-4}

            \cline{2-4}
            
            {}                                                  &
            \multirow{2}{*}{Reset}                              &
            \multicolumn{2}{c?}{Hard}                           \\

            \cline{3-4}

            {}                                                  &
            {}                                                  &
            \multicolumn{2}{c?}{Subtractive}                    \\

            \cline{2-4}

            \cline{2-4}

            {}                                                  &
            \multirow{2}{*}{$\alpha$, $\beta$}                  &
            \multicolumn{2}{c?}{Continuous}                   \\

            \cline{3-4}

            {}                                                  &
            {}                                                  &
            \multicolumn{2}{c?}{$1 - 2^{-n}$}                     \\

            \cline{2-4}

            \cline{2-4}

            {}                                                  &
            $V_{th}$                                            &
            \multicolumn{2}{c?}{Continuous}                     \\

            \cline{2-4}

            \cline{2-4}

            {}                                                  &
            Time-steps                                          &
            \multicolumn{2}{c?}{Set}                            \\

            \hline

            \hline
            
            \multirow{8}{*}{\rotatebox[origin=c]{90}{Training}} &

            Learning rate                                       &
            \multicolumn{2}{c?}{\multirow{4}{*}{Continuous}}    \\

            \cline{2-2}

            \cline{2-2}

            {}                                                  &
            Optimizer                                           &
            \multicolumn{2}{c?}{}                               \\

            \cline{2-2}

            \cline{2-2}

            {}                                                  &
            Regularizer                                         &
            \multicolumn{2}{c?}{}                               \\

            \cline{2-2}

            \cline{2-2}

            {}                                                  &
            Surrogate slope                                     &
            \multicolumn{2}{c?}{}                               \\

            \cline{2-4}

            \cline{2-4}

            {}                                                  &
            Surrogate                                           &
            \multicolumn{2}{c?}{
                \begin{tabular}{c} 
                    Sigmoid, Fast Sigmoid, ATan,    \\
                    Straight Through Estimator,     \\
                    Triangular, SpikeRateEscape,    \\
                    Custom\cite{eshraghian_training_2021}
                \end{tabular}}                                  \\

            \hline
            
            \hline
        \end{tabular}
    \end{table}

\begin{figure*}[t]
    \centering
    \adjustbox{max width=1.2\columnwidth}{
    \begin{lstlisting}[
    basicstyle  = \footnotesize,
    numbers     = left]
        from ax.service.ax_client import AxClient
        
        class SpikExplorer:
        
            def __init__(self, config: dict):
        
                self.num_trials     = config.get("num_trials")
                self.objectives     = config.get("objectives")
                self.search_param   = config.get("search_param")
                self.neurons        = config.get("neurons")
        
            def optimize(self):
        
                dse_engine = AxClient()
        
                # Initialize Bayesian optimization
                dse_engine.create_experiment(
                    parameters=self.search_params,
                    objectives=self.objectives,
                )
        
                search_points = []
        
                for _ in range(self.num_trials):
        
                    # Select initial point in the search space
                    net_config = dse_engine.get_next_trial()
        
                    snn = self.net_generator(net_config)
                    results = self.train_evaluate(snn)
        
                    # Give the results to the optimizer
                    dse_engine.complete_trial(results)
        
                    # Update the Bayesian surrogate model
                    dse_engine.update()
        
                    search_points.append((net_config, results))
        
                return search_points

 \end{lstlisting}
 }

\caption{Summarized code of spike explorer}
 \label{lst:spkexp}
\end{figure*}

As detailed in \autoref{subsec:adse}, Bayesian optimization emerges as the preferred method for \gls{DSE} in \glspl{SNN}. This preference stems from several factors, including the abundance of tunable hyperparameters, inherent noise in the objective function due to the spiking information encoding, and the long training times associated with large \glspl{SNN}. Bayesian optimization is advantageous for its rapid convergence, facilitated by a simplified surrogate model, and its inherently parallelizable nature, accelerating the exploration process.

The \gls{DSE} engine of \spkexp{} is built resorting to the \gls{AX} optimization package, an open-source solution developed at Meta$^{TM}$ \cite{axAdaptiveExperimentation}. It provides high-level \glspl{API} that \spkexp{} uses to iterate through the optimization process efficiently. 
\autoref{lst:spkexp} shows a summarized version of the code used to perform the optimization.
The \gls{DSE} engine receives in input a set of configuration parameters, indicating the number of iterations involved in the optimization (line 7), the objectives of the search (line 8), the metrics to optimize, each associated with the range in which to perform the search (line 9), and the set of candidate neuron architectures, including the functional models and their characterization (line 10).
The optimization process (lines 12-40) starts initializing the Bayesian surrogate model using the \gls{AX} \glspl{API} (lines 14-20) and then performs an iterative procedure (lines 24-38).
A set of parameters is selected at each iteration, following the predictions performed with the surrogate model (line 27). The network is configured with the selected parameters (line 29), and its performance is evaluated (line 30). The results are then provided to the optimizer (line 33), which uses them to update the surrogate model (line 36). The process continues until the required number of iterations is completed (line 24). The full set of explored points (line 38) is returned (line 40). This can be used to find the best configurations on the Pareto frontier and select the configuration that best fits the desired requirements.

\autoref{tab:dse_params} displays the available optimization parameters, organized into three groups: network architecture (net), neuron model (neuron), and training process (training). Numeric parameters are "discrete" or "continuous" ranges. In the former case, only discrete integer values within the specified range are considered, while in the latter case, a continuous interval of real values is analyzed. Additionally, numeric values can be defined as sets of predefined values to try. For non-numerical parameters, enumerative lists of options are used.

Regarding network architecture, \spkexp{} offers constraints for optimizing the model. These constraints include the number of layers to use (discrete range), the number of neurons in each layer (set of options), and the network architecture (feed-forward or recurrent). As exploring the dimensionality of the network is computationally intensive, selecting the number of neurons per layer from a set allows for reducing the search space by performing a coarser search among a predefined range of layer sizes. Conversely, for a finer search, \spkexp{} can be left to select any layer size, and a set containing all integer numbers between the desired minimum and maximum can be provided.
The final parameter related to the network allows for including recurrent connections within layers. This specification occurs at the network level, configuring the entire network with the specified layer type. Hybrid solutions are not currently considered in the search process. As discussed in \autoref{subsec:area}, rather than directly specifying whether layers must be recurrent, users can select models that inherently incorporate recurrence.

At the neuron level, nearly all internal parameters can be adjusted after selecting a specific model among the six options listed in \autoref{tab:dse_params} and elaborated upon in \autoref{subsec:area}. The reset mechanism can be configured as hard or subtractive (refer to \autoref{eq:fire}). Optimization of the exponential decay for both the synaptic current and membrane potential can be achieved through the $\alpha$ and $\beta$ parameters ($0 \leq \alpha \leq 1$ and $0 \leq \beta \leq 1$). In this case, the search can involve continuous values, with users specifying the limits of the search range or selecting from a predefined set of powers of two. The last option aligns with hardware optimization principles, where using powers of two allows for replacing the multiplication involved in exponential decay with a simple bit shift, as demonstrated in \cite{carpegna_spiker_2024}.
Given that exponential decay generally does not require rapid attenuation, $\alpha$ and $\beta$ typically approach values close to one. Consequently, the search primarily focuses on the upper portion of the interval $[0, 1]$, utilizing the expression outlined in \autoref{tab:dse_params}. Additionally, the firing threshold can be adjusted within a continuous range of values to regulate neuron activity. Finally, users can select the number of time steps involved in computation by specifying a set of values. Similar to the approach for selecting the number of neurons, users can limit the set of sequence lengths, tailoring the set's granularity based on the desired search precision. Alternatively, to grant \spkexp{} flexibility in selecting from all possible sequence lengths, users can provide a set containing all integer numbers between the desired minimum and maximum.

In addition to tuning the network architecture, \spkexp{} offers optimization options for the training process. This includes fine-tuning parameters such as the learning rate, optimizer settings —such as the Adam \cite{kingma_adam_2014} parameters $\beta_1$ and $\beta_2$, controlling the decay rates of moving averages of gradients and squared gradients respectively, and influencing the retention of historical information when updating model parameters —regularization parameters like $\lambda$, affecting the strength of L1 and L2 regularization \cite{ng_feature_2004}, and modifying the penalty for large weights, and the surrogate function employed in the backward pass, as elaborated in \autoref{sec:background}. In this scenario, \spkexp{} can select the function itself and adjust its steepness.

Given the many parameters involved, optimization efforts can focus on specific subsets. An illustration of such a targeted search is presented in \autoref{sec:results}.

\section{Experimental results}
\label{sec:results}

This section demonstrates the capabilities of \spkexp{} through selected case studies designed to test its internal optimization engine.

\subsection{Experimental set-up}
\label{subsec:setup}

The exploration capabilities of \spkexp{} were evaluated using three distinct datasets, each with varying complexity and characteristics commonly employed for benchmarking \glspl{SNN}:

    \begin{enumerate}
        \item MNIST~\cite{lecun_gradient-based_1998}: grey-scale images of handwritten digits, converted into sequences of spikes using rate encoding. The corresponding number of inputs is $28 \times 28 = 784$.
        \item \gls{SHD}~\cite{cramer_heidelberg_2022}: audio recordings of numbers pronounced in English and German, converted to spikes through a faithful emulation of the human cochlea. Recordings were performed with $700$ channels, corresponding to the number of inputs of the network.
        \item DVS128~\cite{amir_low_2017}: video recordings of 11 gestures through a \gls{DVS} converting images into  spikes. The sensor's resolution is $128 \times 128$ pixels, accounting for $16,384$ inputs.
    \end{enumerate}

Two optimization experiments were conducted using the three datasets.
In the first experiment, a broad exploration was undertaken, allowing \spkexp{} the freedom to optimize the training process while seeking optimal neuron models and network architectures. The objective was to minimize area and power consumption while maximizing accuracy. The search parameters provided to \spkexp{} are detailed in \autoref{tab:exp_setup}. The exponential decay rates were set to $\alpha = 0.9$ and $\beta = 0.82$ and maintained constant throughout the search process. The number of optimization iterations was selected to constrain the optimization time. It was set to 25 for MNIST and DVS128. Conversely, achieving acceptable accuracy with \gls{SHD} requires more training epochs, so the number of search iterations was capped at 15 to control search duration.
The second experiment focused on a more specific optimization goal. Here, the neuron model was fixed initially, and attention shifted to optimizing the total number of neurons within the network.

Lastly, hardware synthesis of the optimized architecture identified by \spkexp{} for the MNIST dataset was conducted to allow for a comparison with \gls{SOA} \gls{FPGA} accelerators for \glspl{SNN}. The dataset is typically used as the reference benchmark to evaluate \gls{ML} model in general and \gls{SNN} accelerators specifically.

    \begin{table}[h]
            
        \centering

		\caption{Experimental set-up}
		\label{tab:exp_setup}

        \adjustbox{width=\columnwidth}{
        \begin{tabular}{|c|c|c|c|c|c|c|}

            \cline{2-7}

            \multicolumn{1}{c|}{}                       &
            \multicolumn{2}{c|}{\textbf{MNIST}}         &
            \multicolumn{2}{c|}{\textbf{SHD}}           &
            \multicolumn{2}{c|}{\textbf{DVS128}}        \\

            \cline{2-7}

            \multicolumn{1}{c|}{}                       &
            min                                         &
            max                                         &
            min                                         &
            max                                         &
            min                                         &
            max                                         \\
            \hline

            Learning rate                               &
            $10^{-4}$                                   &
            $1.2 \cdot 10^{-4}$                         &
            $10^{-4}$                                   &
            $1.2 \cdot 10^{-4}$                         &
            $10^{-4}$                                   &
            $1.2 \cdot 10^{-4}$                         \\
            
            \hline
            Adam $\beta_{optim}$                        &
            0.9                                         &
            0.999                                       &
            0.9                                         &
            0.999                                       &
            0.9                                         &
            0.999                                       \\
            
            \hline
            \# layers                                   &
            1                                           &
            3                                           &
            1                                           &
            3                                           &
            1                                           &
            3                                           \\

            \hline
            Model                                       &
            \multicolumn{2}{c|}{lif, syn, rlif, rsyn}   &
            \multicolumn{2}{c|}{lif, syn, rlif, rsyn}   &
            \multicolumn{2}{c|}{lif, syn, rlif, rsyn}   \\

            \hline
            Reset                                       &
            \multicolumn{2}{c|}{subtractive}            &
            \multicolumn{2}{c|}{subtractive}            &
            \multicolumn{2}{c|}{subtractive}            \\

            \hline
            Time steps                                  &
            \multicolumn{2}{c|}{10, 25, 50}             &
            \multicolumn{2}{c|}{10, 25, 50}             &
            \multicolumn{2}{c|}{10, 25, 50}             \\

            \hline
            \# neurons/layers                           &
            \multicolumn{2}{c|}{200, 100, 50}           &
            \multicolumn{2}{c|}{200, 100, 50}           &
            \multicolumn{2}{c|}{200, 100, 50}           \\

            \hline
            Search iterations                           &
            \multicolumn{2}{c|}{25}                     &
            \multicolumn{2}{c|}{15}                     &
            \multicolumn{2}{c|}{25}                     \\

            \hline
            Training epochs                             &
            \multicolumn{2}{c|}{50}                     &
            \multicolumn{2}{c|}{100}                    &
            \multicolumn{2}{c|}{50}                     \\

            \hline

        \end{tabular}
        }
        
	\end{table}

\subsection{Global Exploration}
\label{subsec:exploration}

\begin{figure*}[t]
         \centering
         \begin{subfigure}[b]{0.3\textwidth}
             \centering
             \includegraphics[width=\textwidth]{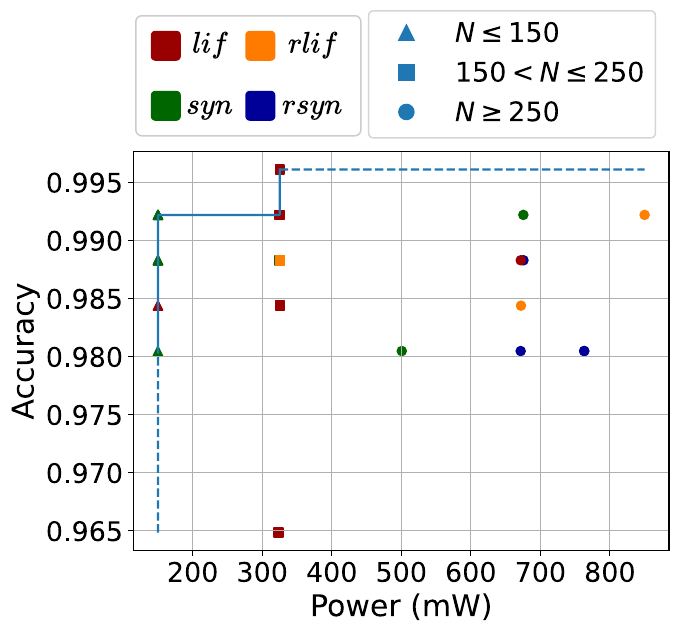}
             \caption{MNIST complete power}
             \label{fig:mnist_complete_power}
         \end{subfigure}
         \hfill
         \begin{subfigure}[b]{0.3\textwidth}
             \centering
             \includegraphics[width=\textwidth]{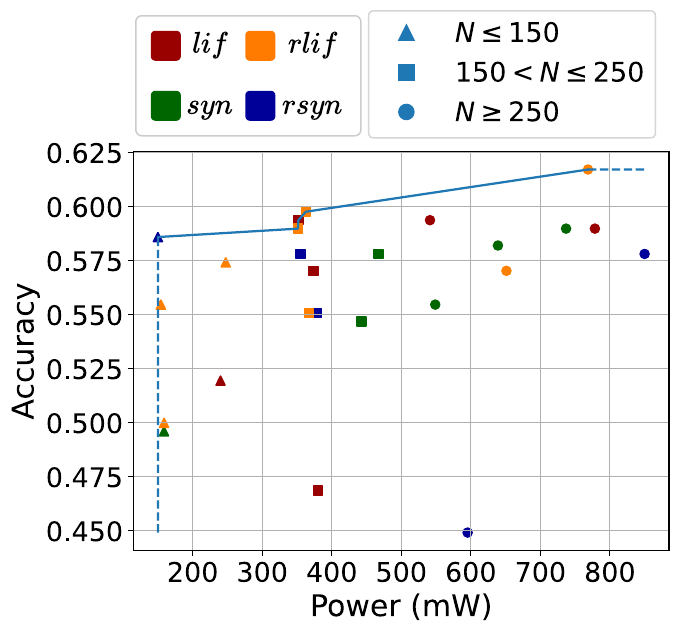}
             \caption{\gls{SHD} complete power}
             \label{fig:shd_complete_power}
         \end{subfigure}
         \hfill
         \begin{subfigure}[b]{0.3\textwidth}
             \centering
             \includegraphics[width=\textwidth]{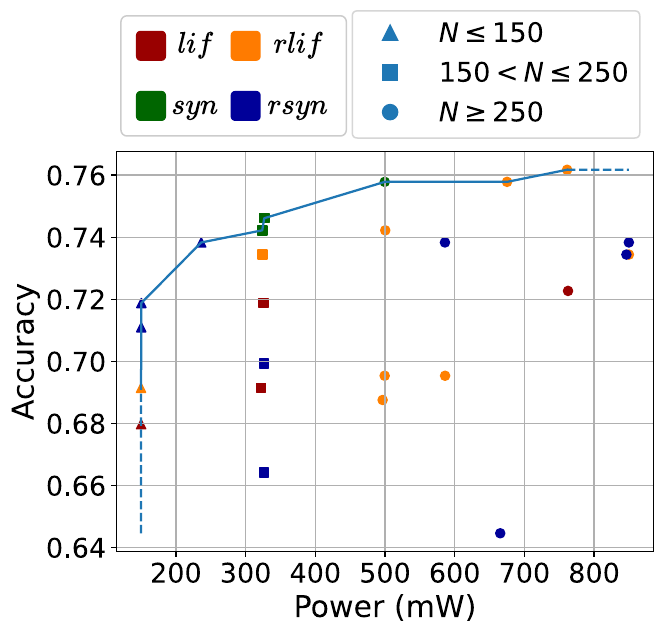}
             \caption{\gls{DVS} complete power}
             \label{fig:dvs_complete_power}
         \end{subfigure}
         \par\bigskip
         \begin{subfigure}[b]{0.3\textwidth}
             \centering
             \includegraphics[width=\textwidth]{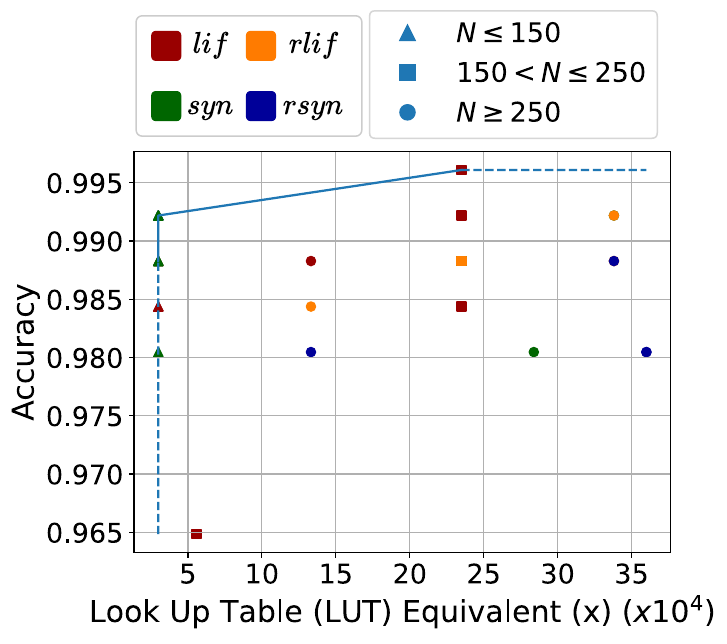}
             \caption{MNIST complete area}
             \label{fig:mnist_complete_area}
         \end{subfigure}
         \hfill
         \begin{subfigure}[b]{0.3\textwidth}
             \centering
             \includegraphics[width=\textwidth]{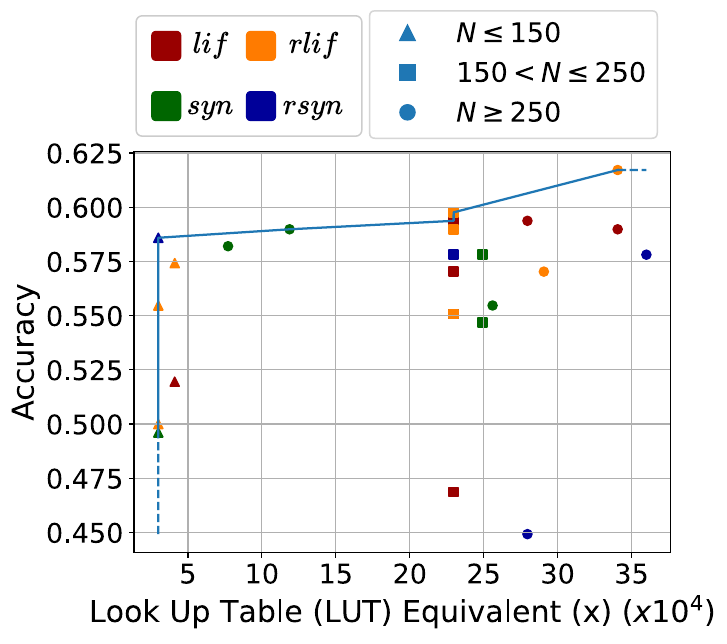}
             \caption{\gls{SHD} complete area}
             \label{fig:shd_complete_area}
         \end{subfigure}
         \hfill
         \begin{subfigure}[b]{0.3\textwidth}
             \centering
             \includegraphics[width=\textwidth]{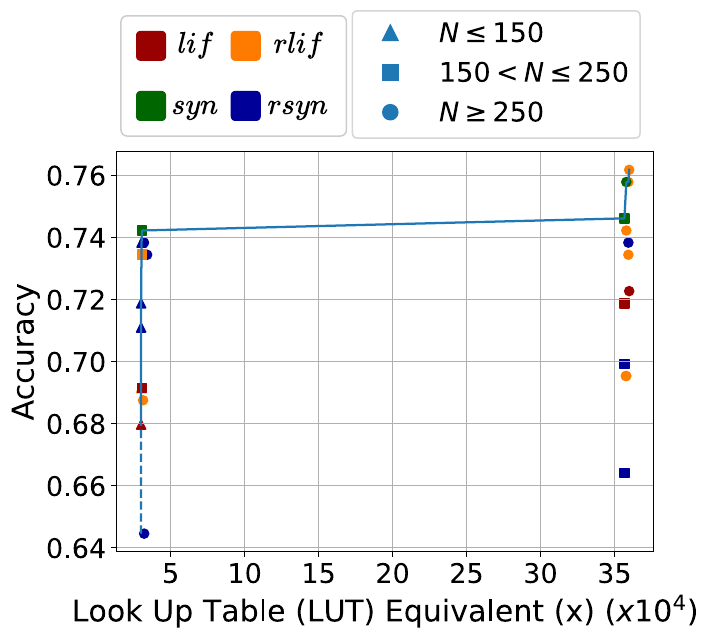}
             \caption{\gls{DVS} complete area}
             \label{fig:dvs_complete_area}
         \end{subfigure}
         \caption{Pareto frontiers of the global exploration on the three benchmark datasets targeting power, area, and accuracy optimization}
         \label{fig:complete_dse}
    \end{figure*}

\autoref{fig:complete_dse} summarizes the performance of \spkexp{} when optimizing the network architecture and parameters for the three selected datasets.
The figure demonstrates a strong correlation between power consumption (the first row in \autoref{fig:complete_dse}) and area (the second row in \autoref{fig:complete_dse}). While this behavior is expected, it is noteworthy because previous publications predominantly emphasized the correlation between power consumption and spiking activity \cite{carpegna_spiker_2024}.
For the MNIST dataset (refer to Figures \ref{fig:mnist_complete_power} and \ref{fig:mnist_complete_area}), non-recurrent models emerge as the preferred choice. This preference is evident from the Pareto frontier, where virtually all top-performing models are \textit{lif} and \textit{syn} without recurrence. Notably, the highest accuracy is achieved with a first-order \gls{LIF} model, devoid of any feedback connection (refer to \autoref{tab:mnist_best_arch}). This is consistent with expectations since simple architectures without explicit recurrent connections should be enough, given the static nature of MNIST data transformed into spike sequences via rate coding.  In this scenario, crucial information is not embedded in the temporal dimension but encoded in the average spike sequence rate. Consistently with what is expected, \spkexp{} converges towards these simpler solutions.
Conversely, in the case of \gls{SHD} and DVS128, acquired through biologically inspired sensors and containing substantial information in spike timing, \spkexp{} generally leans towards recurrent structures such as \textit{rlif} and \textit{rsyn}, along with higher-order models (\textit{syn}). Specifically, for \gls{SHD}, a recurrent structure comprising \textit{rlif} neurons emerges as the favored solution. At the same time, the search tends to diversify more toward both \textit{rsyn} and \textit{rlif}, occasionally incorporating \textit{syn} instances for the DVS128.
It is noteworthy to observe how \spkexp{} can discover superior architectures in terms of accuracy by utilizing the same neuron model and comparable numbers of neurons while playing on other parameters, allowing us to keep the power consumption unchanged while better tuning them on the target task. This is visible when looking at the left section of the Pareto frontier across all three datasets.

In summary, Tables \ref{tab:mnist_best_arch}, \ref{tab:shd_best_arch}, and \ref{tab:dvs_best_arch} showcase the top-1 accuracy optimized \gls{SNN} architecture, parameters, and performance identified by \spkexp{} for the three benchmarks, categorized by neuron model.

\begin{table}[ht!]
        \centering

        \begin{minipage}{0.7\columnwidth}
            
		\caption{Best architectures with the four neuron models on the MNIST}
		\label{tab:mnist_best_arch}

		\begin{adjustbox}{width=\textwidth, center}

			\begin{tabular}{|c|c|c|c|c|}
   
				\hline

                \textbf{Model}		&
				\textbf{Arch.}		&
                \textbf{TS} &
				\textbf{Acc.}	  &
				\begin{tabular}{c}
					\textbf{Power}\\
					(mW)
				\end{tabular}       \\ 

				\hline

                LIF		           & 
				200-10		& 
                10             &
				99.61\%			& 
				310		\\ 

				\hline

                RLIF		           & 
				200-100-200-10		& 
                10             &
				99.22\%			& 
				860		\\ 

				\hline

                SYN		           & 
				200-200-10		&
                25             &
				99.22\%			& 
			    680		\\ 

				\hline

                RSYN		           & 
				100-10		& 
                25             &
				99.22\%			& 
				140		\\ 

				\hline

			\end{tabular}
		\end{adjustbox}

    \end{minipage}
    \par\bigskip
    \begin{minipage}{0.7\columnwidth}
            
		\caption{Best architectures with the four neuron models on the SHD}
		\label{tab:shd_best_arch}

		\begin{adjustbox}{width=\textwidth, center}

			\begin{tabular}{|c|c|c|c|c|}
   
				\hline

                \textbf{Model}		&
				\textbf{Arch.}		&
                \textbf{TS} &
				\textbf{Acc.}		& 
				\begin{tabular}{c}
					Power\\
					(mW)
				\end{tabular}       \\ 

				\hline

                LIF		           & 
				200-20		& 
                50             &
				59.41\%			& 
				360		\\ 

				\hline

                RLIF		           & 
				200-200-20		& 
                50             &
				61.70\%			& 
				760		\\ 

				\hline

                SYN		           & 
				100-100-200-20		&
                10             &
				58.98\%			& 
				720		\\ 

				\hline

                RSYN		           & 
				100-20		& 
                50             &
				58.59\%			& 
				140	\\ 

				\hline

			\end{tabular}
		\end{adjustbox}
    \end{minipage}
    \par\bigskip
    \begin{minipage}{0.7\columnwidth}
            
    \centering

		\caption{Best architectures with the four neuron models on the DVS}
		\label{tab:dvs_best_arch}

		\begin{adjustbox}{width=\textwidth, center}

			\begin{tabular}{|c|c|c|c|c|c|}
   
				\hline

                \textbf{Model}		&
				\textbf{Arch.}		&
                \textbf{TS} &
				\textbf{Acc.}		& 
				\begin{tabular}{c}
					Power\\
					(mW)
				\end{tabular}       \\ 

				\hline

                LIF		           & 
				200-200-50-11		& 
                50             &
				72.27\%			& 
				500		\\ 

				\hline

                RLIF		           & 
				200-200-50-11		& 
                25             &
				76.17\%			& 
				760		\\ 

				\hline

                SYN		           & 
				200-100-11		&
                50             &
				75.78\%			& 
				500			\\ 

				\hline

                RSYN		           & 
				100-200-50-10		& 
                50             &
				73.83\%			& 
				590	\\ 

				\hline

			\end{tabular}

		\end{adjustbox}
    \end{minipage}
    
	\end{table}

To showcase the capability of \spkexp{} across different use cases, this study limited the maximum number of time steps to 50 to mitigate training time. However, upon reviewing the accuracy achieved by the optimized models, it seems reasonable to assume that these datasets may benefit from longer sequences. For instance, \cite{carpegna_spiker_2024} reports a 75\% accuracy for \gls{SHD} with a 200-20 network using \textit{rsyn} neurons and 100-time steps. Conversely, models tailored for MNIST can achieve nearly \gls{SOA} accuracies with minimal time steps.
Regarding architectures, the search for DVS128 tends toward larger structures, which correspondingly increases power consumption.

In terms of computing time, the exploration took approximately 5 hours for both MNIST and \gls{DVS}, and approximately 16 hours for \gls{SHD}, conducted on an AMD Ryzen 9 7950X 16-Core Processor and an Nvidia RTX400 GPU. It is worth noting that the primary time consumption arises from training the network, which is more time-intensive for recurrent models than non-recurrent models due to the inability to accelerate the explicit time dependence through GPU.

\subsection{Fixed neuron models and network size}
\label{subsec:fixed_search}

\begin{figure*}[ht]
         \centering
         \begin{subfigure}[b]{0.3\textwidth}
             \centering
             \includegraphics[width=\textwidth]{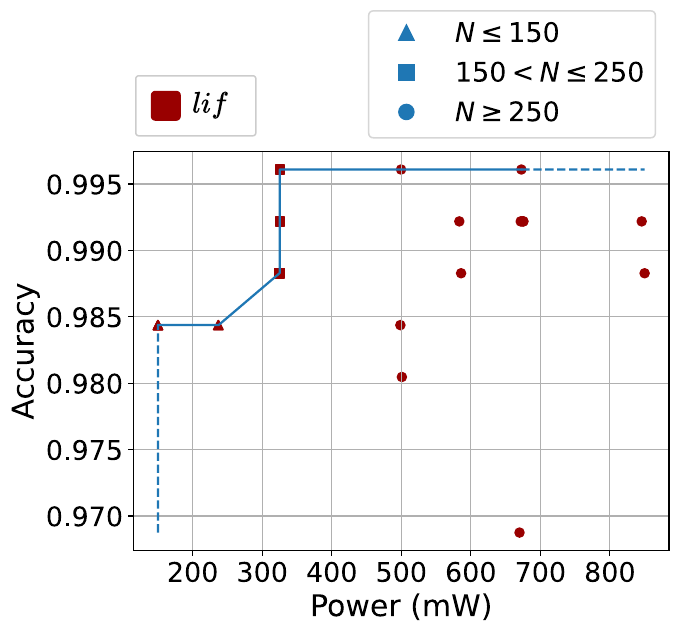}
             \caption{MNIST best model power}
             \label{fig:mnist_best_model_power}
         \end{subfigure}
         \hfill
         \begin{subfigure}[b]{0.3\textwidth}
             \centering
             \includegraphics[width=\textwidth]{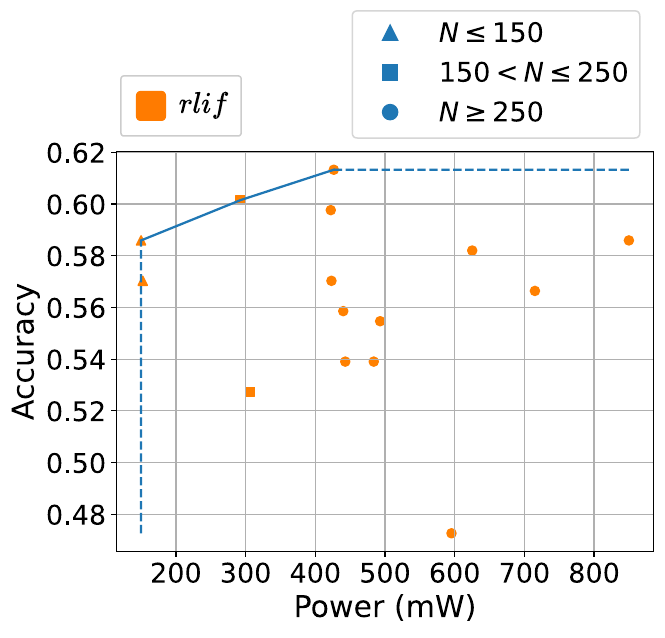}
             \caption{\gls{SHD} best model power}
             \label{fig:shd_best_model_power}
         \end{subfigure}
         \hfill
         \begin{subfigure}[b]{0.3\textwidth}
             \centering
             \includegraphics[width=\textwidth]{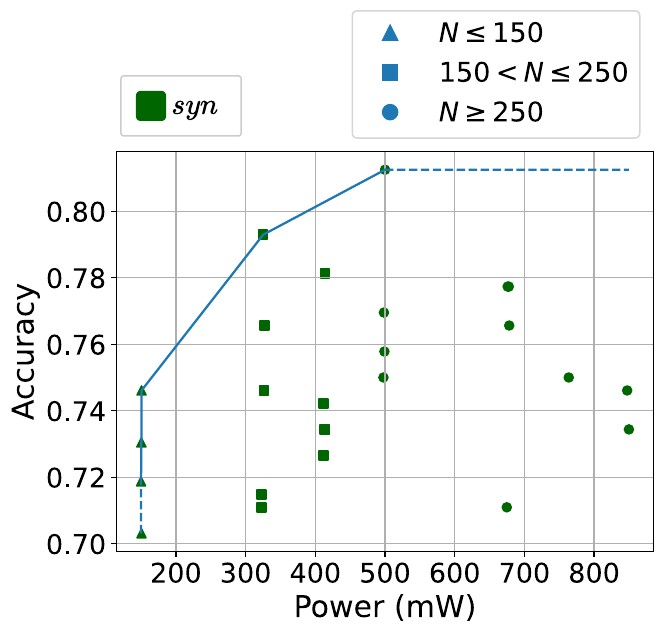}
             \caption{\gls{DVS} best model power}
             \label{fig:fixed models}
         \end{subfigure}
         \caption{Exploration with top accuracy neuron model for each benchmark}
         \label{fig:fixed_models}
    \end{figure*}

    \begin{figure*}[ht]
         \begin{subfigure}[b]{0.3\textwidth}
             \centering
             \includegraphics[width=\textwidth]{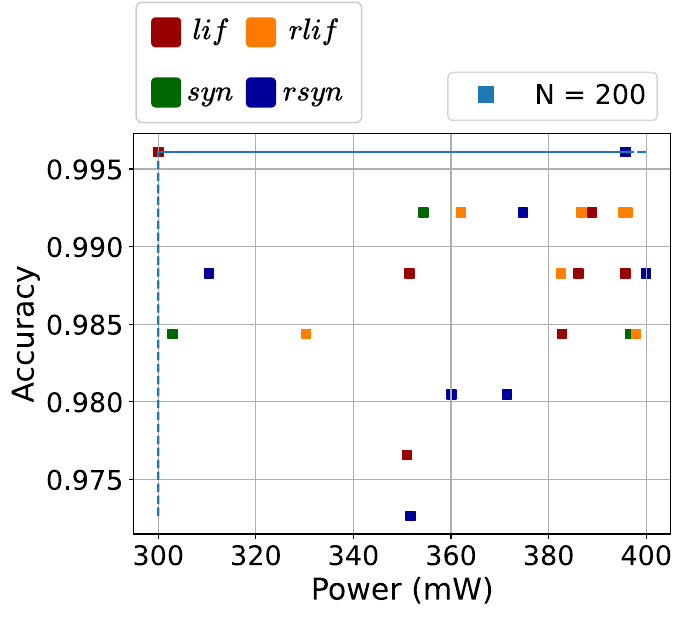}
             \caption{MNIST fixed number of neurons power}
             \label{fig:mnist_best_n_power}
         \end{subfigure}
         \hfill
         \begin{subfigure}[b]{0.3\textwidth}
             \centering
             \includegraphics[width=\textwidth]{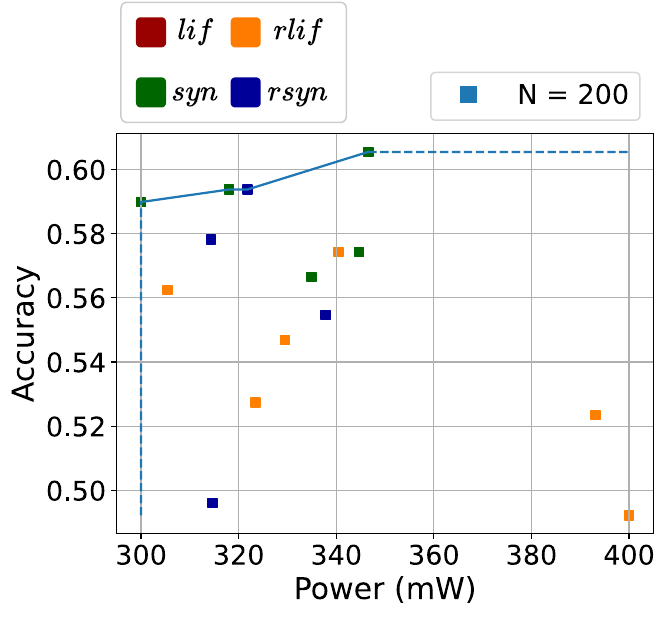}
             \caption{\gls{SHD} fixed number of neurons power}
             \label{fig:shd_best_n_power}
         \end{subfigure}
         \hfill
         \begin{subfigure}[b]{0.3\textwidth}
             \centering
             \includegraphics[width=\textwidth]{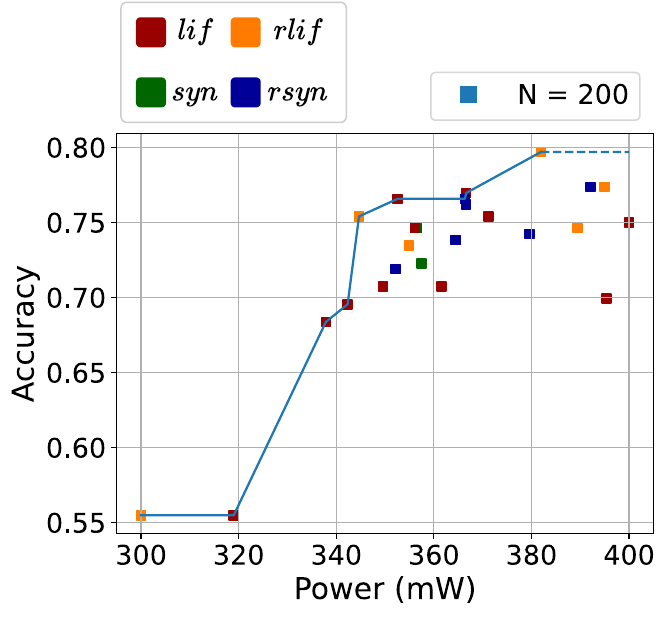}
             \caption{\gls{DVS} fixed number of neurons power}
             \label{fig:dvs_best_n_power}
         \end{subfigure}
         \caption{Exploration with number of neurons constrained to 200 for each benchmark}
         \label{fig:fixed_neurons}
    \end{figure*}

After showcasing the overall optimization capabilities of \spkexp{}, additional experiments were performed to highlight its behavior in constrained optimization problems. 

\autoref{fig:fixed_models} showcases the capability of \spkexp{} to optimize the network with a predefined neuron model, solely using the network architecture and parameters. In this case, the Pareto frontiers are dominated by small architectures ($N \le 250$). It is interesting to observe that the optimizer is very effective when selecting the network architecture: for example, for the MNIST, an architecture with 200-10 neurons (square on the top left of Pareto frontier) can obtain the same accuracy of bigger solutions (circles on the top right of the Pareto frontier), reducing the power requirements by factors of 1.5 and more than 2, respectively. Results reported in \autoref{fig:fixed_models} also highlight the capability of \spkexp{} in supporting designers in finding the suitable trade-off between different metrics. For example, the Pareto frontier in \autoref{fig:shd_best_model_power} shows that the power can be reduced almost 3 times by accepting an accuracy loss of 3\%.
Interestingly, the accuracy on the DVS128 is pushed up to the best value of 81.6\%, improving by around 5\% concerning a more agnostic search, indicating that a more specialized search can reach even better results.
    
Finally, \autoref{fig:fixed_neurons} shows the behavior of \spkexp{} when constraining the total number of neurons to 200 to study how different models behave. Interestingly, this produces different observations compared to results reported in \autoref{subsec:exploration}. The optimization for \gls{SHD} privileges now the \textit{syn} model, either with or without recurrent connections, while the \textit{lif} model dominates the Pareto frontier in the DVS128 case. Again, the top-1 accuracy is increased, even if less than in the search with a fixed neuron model, reaching around 80\%. This again supports the utility of a tool like \spkexp{} when exploring different design opportunities.

\subsection{Synthesis and comparison with State of Art}
\label{subsec:synthesis}

As discussed in \autoref{sec:methods}, the power and area values provided by \spkexp{} are estimations to guide the \gls{DSE} process and do not represent the actual values of the final \gls{FPGA} implementation of the respective model.
To obtain actual values and compare the performance of the models optimized by \spkexp{} with \gls{SOA} \gls{SNN} accelerators designed for \gls{FPGA}, an effective synthesis of an optimized architecture was conducted using the Spiker+ framework to generate the VHDL description \cite{carpegna_spiker_2024}. This process generated the hardware implementation of a 128-10 architecture optimized with \spkexp{} for the MNIST dataset.
This architecture is compared with other accelerators in \autoref{tab:mnist_comparison}.

An important observation is that the same architecture, with the same neuron model used in \cite{carpegna_spiker_2024}, is considered for a direct comparison. All other parameters are optimized following the approach outlined in subsections \ref{subsec:exploration} and \ref{subsec:fixed_search}.
In this scenario, \spkexp{} optimizes the model by reducing the time steps from 100 to 16, decreasing the overall latency by more than 6 times, from $780\mu s$ to $120\mu s$. Simultaneously, the optimized training increases the accuracy by almost 3\%, reaching 95.8\%, thereby establishing the new optimized model as the best one among those considered, both in terms of power consumption and latency, while also positioning it close to the best-performing model in terms of accuracy \cite{han_hardware_2020}. Thus, \spkexp{} demonstrates its capability to enhance the design of \gls{FPGA} accelerators for \glspl{SNN}, simplifying the selection of the optimal architecture and effectively tailoring it to the desired application.

	\begin{table*}[t]

		\caption{Comparison of \spkexp{} to state-of-the-art \gls{FPGA} accelerators for \glspl{SNN}}
		\label{tab:mnist_comparison}

        \centering

		\begin{adjustbox}{width=0.8\textwidth, center}

			\begin{tabular}{|c|c|c|c|c|c|c|c|}

				\hline

				\textbf{Design}				& 
				Han et al.\cite{han_hardware_2020}	&
				Gupta et al. \cite{gupta_fpga_2020}	&
				Li et al.\cite{li_fast_2021}		&
				SPIKER\cite{carpegna_spiker_2022}				&
				SPIKER+ \cite{carpegna_spiker_2024} &
                This work                      \\

				\hline

				\textbf{Year}				& 
				2020					&
				2020					&
				2021					&
				2022					&
				\multicolumn{2}{c|}{2024}\\

				\hline

				\textbf{$f_{clk}$[MHz]}			& 
				200					&
				100					&
				100					&
				\multicolumn{3}{c|}{100}  \\

				\hline

				\textbf{Neuron bw}			&
				16					&
				24					&
				16					&
				16					&
				\multicolumn{2}{c|}{6}  \\

				\hline

				\textbf{Weights bw}			&
				16					&
				24					&
				16					&
				16					&
				\multicolumn{2}{c|}{4}  \\

				\hline

				\textbf{Update}				&
				Event					&
				Event					&
				Hybrid					&
				\multicolumn{3}{c|}{Clock}  \\

				\hline

				\textbf{Model}				&
				LIF					&
				LIF\cite{iakymchuk_simplified_2015}	&
				LIF					&
				\multicolumn{3}{c|}{LIF}  \\

				\hline
				\textbf{FPGA}				&
				XC7Z045					&
				XC6VLX240T				&
				XC7VX485			&
				\multicolumn{3}{c|}{XC7Z020}\\

				\hline

				\textbf{Avail. BRAM}			&
				545					&
				416					&
				2,060					&
				\multicolumn{3}{c|}{140}  \\

				\hline

				\textbf{Used BRAM}			&
				40.5					&
				162					&
				N/R					&
				45					&
				\multicolumn{2}{c|}{18}       \\

				\hline

				\textbf{Avail. DSP}			&
				900					&
				768					&
				2,800					&
				\multicolumn{3}{c|}{220}      \\

				\hline

				\textbf{Used DSP}			&
				0					&
				64					&
				N/R					&
				\multicolumn{3}{c|}{0}  \\

				\hline

				\textbf{Avail. logic cells}		&
				655,800					&
				452,160					&
				485,760					&
				\multicolumn{3}{c|}{159,600}    \\

				\hline

				\textbf{Used logic cells}		&
				12,690					&
				79,468					&
				N/R					&
				55, 998					&
				\multicolumn{2}{c|}{7,612}         \\

				\hline

				\textbf{Arch}				&
				1024-1024-10			&
				784-16					&
				200-100-10				&
				400					&
				\multicolumn{2}{c|}{128-10}      \\

				\hline

				\textbf{\#syn}				&
				1,861,632				&
				12,544					&
				177,800					&
				313,600					&
				\multicolumn{2}{c|}{101,632}		 \\

				\hline

				\textbf{$T_{lat}$/img [ms]}		&
				6.21					&
				0.50					&
				3.15					&
				0.22					&
        		0.78                    &
                0.12      \\

				\hline

				\textbf{Power [W]}			&
				0.477					&
				N/R					&
				1.6					&
				59.09					&
				\multicolumn{2}{c|}{0.18}                   \\

				\hline

				\textbf{E/img [$mJ$]}			&
				2.96					&
				N/R					&
				5.04					&
				13					&
				0.14                &
                0.02\\

				\hline

				\textbf{E/syn [$nJ$]}&
				1.59					&
				N/R					&
				28					&
				41					&
                1.37                &
                0.22                \\

				\hline



				\textbf{Accuracy}			&
				97.06\%					&
				N/R					&
				92.93\%					&
				73.96\%					&
				93.85\%		&
                95.8\%      \\

				\hline

			\end{tabular}

		\end{adjustbox}
	\end{table*}

 \section{Conclusions}
\label{sec:discussion}


This paper introduced \spkexp{}, a tool tailored for hardware-centric \gls{ADSE} in \glspl{SNN}. Specifically designed for crafting and fine-tuning specialized hardware accelerators intended for deployment on \gls{FPGA}, this tool showcases the effectiveness of Bayesian optimization within the context of \glspl{SNN}. It enables an easy and flexible multi-objective search, considering model accuracy and critical hardware-specific metrics such as power consumption, area utilization, and latency. The design of \spkexp{} builds upon three open-source projects: snnTorch, \gls{AX}, and SPIKER+. Being open-source, \spkexp{} offers a robust solution for optimizing \glspl{SNN}.

The capabilities of \spkexp{} were evaluated across three distinct tasks: static image recognition using the MNIST dataset, a prevalent benchmark in \gls{ML}; speech recognition on the \gls{SHD} dataset; and gesture recognition on the DVS128 dataset.
In the MNIST scenario, the tool achieved outstanding performance, surpassing existing solutions in terms of latency by classifying images in approximately $120\mu s$, while consuming minimal power (180mW) and achieving high accuracy (95.8\%).
On the \gls{SHD} task, it encountered challenges, achieving a top-1 accuracy of approximately 62\%, possibly due to the limited number of time steps used for spike sequences during optimization.
Regarding the DVS128 dataset, \spkexp{} delivered promising results, achieving an 81.6\% top-1 accuracy. Notably, the high dimensionality of the inputs of this dataset, with 128x128 event-based channels, made the use of \gls{FC} networks suboptimal and fully parallel processing infeasible. Nevertheless, this dataset served as a valuable case study for evaluating the optimization tool with a complex dataset.

Future work involves expanding the framework's scope to encompass different architectures such as \gls{CSNN} and generalizing the tool to accommodate diverse computing paradigms like event-driven processors. Despite not being explicitly tailored for such hardware accelerators, \spkexp{} exhibits considerable flexibility, supporting custom neuron models and configurable metric assessments during optimization. This lays a solid foundation for automating the optimization of \gls{SNN} co-processors, thereby facilitating the adoption of neuromorphic solutions in resource-constrained edge applications.

\bibliographystyle{plain}
\bibliography{bibliography}

\end{document}